\documentclass[10pt,twocolumn,letterpaper]{article}

\usepackage[preprint]{cvpr}      
\definecolor{cvprblue}{rgb}{0.21,0.49,0.74}
\usepackage[pagebackref,breaklinks,colorlinks,allcolors=cvprblue]{hyperref}

\def\paperID{19373} 
\def\confName{CVPR}
\def\confYear{2026}

\title{Dummy-Aware Weighted Attack (DAWA): Breaking the Safe Sink in Dummy Class Defenses}

\author{Yunrui Yu\\
Tsinghua University\\
{\tt\small yuyunrui@mail.tsinghua.edu.cn}
\and
Xuxiang Feng\\
University of Macau\\
{\tt\small fengxx@aircas.ac.cn}
\and 
Pengda Qin \\
{\tt\small derelpdqin@tencent.com} 
\and
Pengyang Wang \\
University of Macau\\
{\tt\small pywang@um.edu.mo}
\and
Kafeng Wang \\
  Tsinghua University \\
{\tt\small wangkafeng@mail.tsinghua.edu.cn} 
\and 
Cheng-zhong Xu \\
  University of Macau \\
{\tt\small chengzhongxu@um.edu.mo} 
\and
Hang Su\\
Tsinghua University\\
{\tt\small suhangss@mail.tsinghua.edu.cn}
\and
Jun Zhu\\
Tsinghua University\\
{\tt\small dcszj@mail.tsinghua.edu.cn}
}

\usepackage{tikz}
\usetikzlibrary{arrows.meta, positioning, shapes.misc}
\usepackage{mathtools}
\usepackage{amssymb}
\usepackage{amsmath}
\usepackage{bm,yhmath}
\usepackage{nicefrac} 
\usepackage{algorithm,algpseudocode}
\usepackage{adjustbox,multirow,siunitx}
\DeclarePairedDelimiter{\parens}{\lparen}{\rparen}

\DeclarePairedDelimiter{\bracks}{[}{]}

\newcommand{\inputset}{\mathcal{I}}
\newcommand\attack{{DAWA}}
\newcommand\attackmt{{DAWA}\textsuperscript{mt}}

\newcommand{\cifarx}{CIFAR-10}
\newcommand{\cifarc}{CIFAR-100}

\newcommand{\textlnorm}[1]{\( \ell_{#1} \)-norm}

\newcommand{\uniform}[1]{\mathcal{U}\left({#1}\right)}
\newcommand{\realset}{\mathbb{R}}

\newcommand{\x}{\mathbf{x}}

\newcommand{\advset}{\mathcal{A}_{\epsilon, \x}}
\newcommand{\xadv}{\hat\x}
\newcommand{\y}{y}

\newcommand{\f}[1]{{f}_{#1}}

\newcommand{\m}{\bm{\mu}}

\newcommand{\outputset}{\realset^K}
\newcommand{\stepsize}{\beta}
\newcommand{\momentum}{\nu}
\newcommand{\tbgreen}[1]{\textcolor[rgb]{0,0.502,0}{#1}}

\newcommand{\aaa}{\( \mathrm{A}^3 \)}

\DeclareMathOperator{\loss}{\mathcal{L}}

\DeclareMathOperator{\project}{%
    \mathcal{P}_{\epsilon, \x}}
\DeclareMathOperator{\sign}{\mathrm{sign}}

\DeclareMathOperator{\attackloss}{%
    \mathcal{L}^\mathrm{dawa}}

\makeatletter%
\newcolumntype{L}{D{.}{.}{2,2}}
\newcolumntype{B}[3]{>{\boldmath\DC@{#1}{#2}{#3}}c<{\DC@end}}
\newcommand{\tbox}[1]{\begin{tabular}[c]{@{}c@{}}#1\end{tabular}}

\begin{document}
\maketitle
\begin{abstract}
Adversarial robustness evaluation faces a critical challenge as new defense paradigms emerge that can exploit limitations in existing assessment methods. This paper reveals that Dummy Classes-based defenses, which introduce an additional ``dummy'' class as a safety sink for adversarial examples, achieve significantly overestimated robustness under conventional evaluation strategies like AutoAttack. The fundamental limitation stems from these attacks' singular focus on misleading the true class label, which aligns perfectly with the defense mechanism—successful attacks are simply captured by the dummy class. To address this gap, we propose Dummy-Aware Weighted Attack (DAWA), a novel evaluation method that simultaneously targets both the true label and dummy label with adaptive weighting during adversarial example synthesis. Extensive experiments demonstrate that DAWA effectively breaks this defense paradigm, reducing the measured robustness of a leading Dummy Classes-based defense from \textbf{58.61\%} to \textbf{29.52\%} on CIFAR-10 under $\ell_\infty$ perturbation ($\epsilon=8/255$). Our work provides a more reliable benchmark for evaluating this emerging class of defenses and highlights the need for continuous evolution of robustness assessment methodologies.
\end{abstract}    
\section{Introduction}\label{sec:intro}

Deep neural networks (DNNs) have witnessed remarkable achievements in various safety-critical domains, including autonomous robots~\cite{filos2020can,vulpi2021recurrent}, self-driving vehicles~\cite{eykholt2018robust}, and search engines~\cite{tolias2019targeted}. However, the increasing reliance on DNNs in these applications also exposes them to a substantial vulnerability: adversarial examples. These examples involve the addition of minuscule intentional perturbations to the input data, leading the model to produce erroneous outputs~\cite{szegedy14,goodfellow15}. Consequently, the emergence of adversarial attacks poses a significant threat to the safety and dependability of deep learning applications.

To address this critical threat, the research community has witnessed an ongoing arms race between the development of defense strategies and the advancement of attack methodologies for robustness evaluation. On the defense front, numerous approaches~\cite{goodfellow15,moosavi2016deepfool,carlini17, madry18,dong18momentum,xiao2019advgan,croce20aa} have been proposed to enhance model robustness, with adversarial training~\cite{madry18,zhang19trades,carmon19} emerging as one of the most effective paradigms. However, it introduces a well-documented trade-off: while improving robustness against adversarial examples, it often compromises accuracy on clean, natural images~\cite{tsipras2018robustness}. 

\begin{figure}[ht]
    \centering%
    \includegraphics[width=\linewidth, trim=0 10pt 0 0]{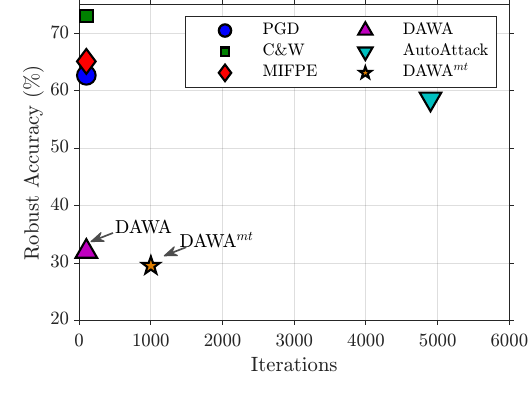}
    \caption{%
        Attack effectiveness comparison on CIFAR-10 with $\ell_\infty$ constraint under PGD-AT+DUCAT defense ~\cite{wang2024new}. Our DAWA (100 iterations, non-targeted) and DAWA$^{mt}$ (1,000 iterations, combined non-targeted/targeted) achieve $35.60\%$ and $26.42\%$ robust accuracy respectively, outperforming PGD ($60.64\%$), C\&W ($71.72\%$), MIFPE ($63.10\%$), and AutoAttack ($56.80\%$).
        }\label{fig:introduction:attack_com}
\end{figure}

Concurrently, a variety of attack methods have been developed to rigorously evaluate these defenses~\cite{madry18,shafahi19free,alayrac19,zhang19trades,pang20hypersphere,wang20misclass,wu20wp,wu20width,lafeat,gao2022mora,yu2023lafit}. Projected Gradient Descent (PGD)~\cite{madry18} was once widely adopted as a standard benchmark. Nevertheless, studies soon revealed that gradient-based attacks like PGD could significantly overestimate model robustness, a failure often attributed to \textit{gradient masking}~\cite{goodfellow2018gradient}. To overcome this limitation, subsequent research pursued more reliable evaluation strategies. AutoAttack~\cite{croce20aa}, and Composite Adversarial Attacks (CAA)~\cite{mao2021caa}, which ensemble diverse attack algorithms, were proposed to mitigate overestimation, though at the cost of substantial computational overhead.

Parallel to these empirical advances, other work has delved into the root causes of the overestimation problem. The study by~\cite{yu2023efficient} identified that the relative error in gradients, exacerbated by floating-point arithmetic, is a fundamental factor leading to robustness overestimation by gradient-based attacks. To counter this, they proposed the MIFPE loss function, which effectively mitigates the negative impact of numerical errors. Their method demonstrated remarkable efficiency, achieving results comparable to AutoAttack's 4900 iterations in just 100 iterations for untargeted attacks. Furthermore, the theoretical analysis in~\cite{yu2025theoretical} revealed that the hyperparameter value of 1 used in MIFPE is very close to the theoretical optimum, providing a mathematical explanation for its exceptional performance.

Despite these advancements, a fundamental challenge remains: evaluation strategies are often designed with existing defenses in mind, making them inherently susceptible to being outpaced by novel defense paradigms. As illustrated in \Cref{fig:introduction:attack_com}, this is precisely the case with the recently proposed \textbf{Dummy Classes-based defenses}~\cite{wang2024new}.  These defenses achieve state-of-the-art reported robustness under AutoAttack evaluation, but this robustness is severely overestimated. Their core innovation lies in introducing an additional, synthetically created ``dummy'' class and \textbf{releasing the inherent strong constraint} of traditional adversarial training. Instead of forcing adversarial examples to be classified as their original true class, they are explicitly guided toward the dummy class during training. This creates a "safe sink" that neutralizes conventional attacks, including AutoAttack, which focus solely on misleading the true class label—a critical flaw that this new defense strategically exploits: when a conventional attack successfully reduces the probability of the true class, the gradient-based optimization naturally pushes the example into the dummy class—the very outcome the defense is trained to achieve. Consequently, these defenses can deliberately exploit the weaknesses of traditional evaluation strategies, leading to a severe overestimation of their robustness.

The root cause of this overestimation lies in the mismatch of objectives. A successful attack must simultaneously achieve two goals: lead the model away from the true class label, and actively pull the example away from the dummy class sink. Attacking only the ensemble output as a single entity is insufficient because the loss function is not explicitly designed to avoid this trap. This necessitates an attack strategy that explicitly and simultaneously targets both the authentic and virtual labels, dynamically weighting their importance during the perturbation process to find a path to a different, real class.

Based on the observations above, it becomes evident that a reliable evaluation of defenses incorporating dummy classes requires a fundamental shift in attack methodology. The root cause of AutoAttack's overestimation lies in the fundamental mismatch of objectives: while a successful attack must simultaneously lead the model away from both the true class label and the dummy class sink, conventional attacks focus solely on misleading the true class label. This critical limitation necessitates an attack strategy that explicitly and simultaneously targets both the authentic and virtual labels. To address this gap, this paper introduces \textbf{DAWA (Dummy-Aware Weighted Attack)}. DAWA is specifically designed to break the ``safe sink'' illusion by simultaneously attacking the true label and the dummy label, while adaptively re-weighting their contributions during adversarial example synthesis. This ensures the generated perturbations effectively navigate away from both the original class and the defensive dummy class, forcing the model into making a meaningful error on another real class.

Extensive experiments demonstrate that DAWA drastically reduces the overstated robustness of Dummy Classes-based defenses. For instance, on CIFAR-10 under \( \ell_\infty \) perturbation (\( \epsilon = 8/255 \)), a leading Dummy Classes-based defense that AutoAttack certifies at \textbf{58.61\%} robustness drops to only \textbf{29.52\%} against DAWA. By pushing the boundaries of current SOTA evaluation techniques specifically for this emerging class of defenses, DAWA provides a more nuanced, adaptive, and reliable benchmark for assessing the adversarial robustness of Dummy Classes-based defenses.

We summarize our contributions as follows:
\begin{itemize}
    \item We reveal the fundamental reason why AutoAttack overestimates the robustness of Dummy Classes-based defenses: its singular focus on misleading the true class label, which aligns perfectly with the defense's ``safe sink'' mechanism.
    
    \item We propose \textbf{DAWA}, a novel attack that simultaneously targets both the true label and the dummy label with adaptive weighting, effectively breaking the defense's protection mechanism.
    
    \item Extensive experiments show DAWA reduces the measured robustness of a leading defense from 58.61\% to 29.52\% on CIFAR-10, providing a more reliable evaluation benchmark for this emerging defense paradigm.
\end{itemize}
\section{Related Work}

\subsection{Adversarial Attacks}

The field of adversarial machine learning has matured significantly with the establishment of reliable, standardized evaluation benchmarks. Early attacks such as the Fast Gradient Sign Method (FGSM)~\cite{goodfellow15} and its iterative variant, Projected Gradient Descent (PGD)~\cite{madry18}, laid the foundational paradigm for crafting adversarial examples. Formally, given a classifier \( f: \inputset \to \outputset \) and an input \( \x \) with true label \( y \), an adversarial example \( \xadv \) is sought within an \( \ell^p \)-norm ball \( \advset \) such that:
\begin{equation}
    \arg\max f\parens{\xadv} \neq y.
    \label{eq:adversarial_example}
\end{equation}
The PGD attack iteratively maximizes a classification loss (typically the softmax cross-entropy) to achieve this:
\begin{equation}
    \xadv_{i + 1} = \project\parens{
        \xadv_i + \alpha_i \sign\parens{
            \nabla \loss\parens{f\parens{\xadv_i}, y}
        }
    },
\end{equation}
where
\( \loss \) is typically
the softmax cross-entropy (SCE) loss
used to train the model,
 \( \alpha_i \) is the step size,
and we let the initial
\( \xadv_0 \triangleq \project\parens{\x + \m} \).
The projection function \( \project\parens{\mathbf{v}} \)
constrains its input \( \mathbf{v} \)
to be within the feasible region \( \advset \),
and finally
\( \m \sim \uniform{-\epsilon, \epsilon} \)
is a uniformly distributed noise
bounded by \( \bracks{-\epsilon, \epsilon} \).

However, the proliferation of diverse defense mechanisms led to an "arms race" and an evaluation crisis, where many defenses were later found to rely on "gradient masking" or obfuscation, leading to inflated and non-comparable robustness claims. This highlighted the critical need for a rigorous, standardized, and reliable evaluation benchmark.

The introduction of \emph{AutoAttack (AA)}~\cite{croce20aa} marked a pivotal moment in addressing this challenge. As a parameter-free, ensemble-based attack that combines diverse strategies, AA quickly became the \emph{de facto} gold standard for robustness evaluation. Its core objective---ensuring that the logit of the true class \( z_y \) is less than the maximum logit of any other class \( \max_{i \neq y} z_i \)---became the canonical criterion for a successful attack. This objective provided a unified and powerful framework that significantly elevated the rigor of the field. Subsequent works like LAFEAT~\cite{lafeat} and Adaptive Auto Attack (\aaa)~\cite{ye2022aaa} further strengthened the attack arsenal, while Composite Adversarial Attacks (CAA)~\cite{mao2021caa} automated the selection of effective attack sequences. The community had thus converged on a robust evaluation paradigm centered on powerful, standardized attacks.

Parallel to these empirical advances, other work has delved into the root causes of the overestimation problem. The study by~\cite{yu2023efficient} identified that the relative error in gradients, exacerbated by floating-point arithmetic, is a fundamental factor leading to robustness overestimation by gradient-based attacks. To counter this, they proposed the MIFPE loss function:
\begin{equation}
    \mathcal{L}^{\text{MIFPE}}\left( \mathbf{z}, y \right) \triangleq \mathcal{L}^{\text{ce}}\left( \frac{t^* \mathbf{z}}{(\mathbf{z}_{\pi_1}- \mathbf{z}_{\pi_2})_{\text{detach}}}, y \right),
\end{equation}
which effectively mitigates the negative impact of numerical errors. Their method demonstrated remarkable efficiency, achieving results comparable to AutoAttack's 4900 iterations in just 100 iterations for untargeted attacks. Furthermore, the theoretical analysis in~\cite{yu2025theoretical} revealed that the hyperparameter value of \( t^* = 1 \) used in MIFPE is very close to the theoretical optimum, providing a mathematical explanation for its exceptional performance.

\subsection{Adaptive Defenses Exploiting the Evaluation Paradigm}

In response to these powerful and standardized attacks, the defense community evolved. While early defenses like \emph{adversarial training}~\cite{madry18} aimed to directly solve the saddle-point problem by minimizing loss on adversarial examples, a new class of defenses emerged that strategically exploited the very assumptions underpinning the evaluation benchmarks.

These defenses can be broadly categorized into several paradigms. \emph{Adversarial Training} and its variants~\cite{madry18,zhang19trades,wu20wp} directly optimize the model parameters to be robust against adversarial perturbations by incorporating adversarial examples during training. \emph{Input Preprocessing} defenses~\cite{liao2018defense,zhang2019defense,nie2022diffusion} aim to "purify" or remove adversarial perturbations from the input before feeding it to the classifier, often using techniques like denoising or generative models. \emph{Randomization-based} approaches~\cite{xie2017mitigating,raghunathan2018certified} introduce stochasticity either at inference time or during model components to break the deterministic gradient calculations that attacks rely on.  \emph{Certifiably robust} methods~\cite{cohen2019certified,salman2019provably,gowal2020uncovering} have emerged that provide mathematical guarantees of robustness within a certain perturbation bound. More recently, a particularly insightful strategy involves \emph{architectural modifications} that intentionally alter the model's output space.
A seminal and representative work in this direction is the \emph{Dummy Classes-based} defense paradigm~\cite{wang2024new}.

The core mechanism of these defenses is to "dummy" classes. Formally, a classifier \( f \) originally mapping to \( K \) classes is modified to output logits for \( 2K \) classes, where \( K+1,..., 2K \)-th class is the dummy class. The training objective is ingeniously designed to:
1) Classify clean samples to their true label \( y \),
2) Explicitly guide adversarial examples \( \xadv \) to be predicted as the dummy class \( K+y \), and
3) Actively encourage a discrepancy between the predictions for \( \x \) and \( \xadv \).

The critical insight is that these defenses are explicitly designed to \emph{satisfy} the success criterion of traditional attacks like AutoAttack. When an attack successfully reduces the logit of \( y \) below that of the dummy class \( K+y \) (i.e., \( z_y < z_{K+y} \)), AutoAttack truthfully reports a successful attack (\( \arg\max f(\xadv) = K+y \neq y \)). However, this is an illusion of robustness. The defense mechanism can simply \emph{map the dummy class prediction back to the original true class \( y \)} during inference, thereby nullifying the attack from a functional perspective. This creates a "safe sink" that artificially inflates the measured robustness against traditional evaluation benchmarks.

\subsection{The Inadequacy of Conventional Attacks Against Adaptive Defenses}

The emergence of Dummy Classes-based defenses exposes a fundamental limitation in the prevailing evaluation paradigm. Attacks like AutoAttack, C\&W, and their variants, which singularly focus on the objective \( \arg\max f(\xadv) \neq y \), are inherently ill-suited to assess the true robustness of these adaptive defenses.

Their failure stems from a critical \emph{misalignment of objectives}. The traditional attacker's goal is simply to make \( \max_{i \neq y} z_i \) greater than \( z_y \), and the defense happily complies by offering the dummy class as the "easiest" \( i \neq y \) to maximize. The attacker, following the standard protocol, declares victory, while the defender, in practice, remains unharmed. This leads to a severe \emph{overestimation} of the defense's true robustness.

While subsequent works have proposed more powerful attacks, such as adaptive attacks~\cite{tramer2020adaptive} that manually tailor strategies for specific defenses, they often require significant manual effort and insight for each new defense. More importantly, they do not address the \emph{systemic flaw} in the core objective function shared by most automated evaluation tools. The community lacks a generalized attack framework that can automatically and reliably evaluate defenses that exploit this objective function loophole.

\subsection{Our Work: Towards a Next-Generation Robustness Evaluation}

The limitations discussed above underscore the necessity for a new evaluation paradigm. To accurately assess the robustness of modern adaptive defenses like those based on dummy classes, an attack must look beyond the simplistic "y vs. non-y" dichotomy.

In this paper, we bridge this critical gap. We propose \textbf{DAWA}, a novel attack framework designed from the ground up to counter defenses that exploit the traditional evaluation objective. The core innovation of our method is a refined objective function that \emph{simultaneously considers the true class \( y \) and the defender's designated safe sink (the dummy class)}. Our attack does not merely seek to deviate from \( y \); it actively works to prevent the adversarial example from being captured by the dummy recovery mechanism. By doing so, \textbf{DAWA} provides a more accurate, reliable, and truthful estimation of the model's real-world robustness, moving the evaluation benchmark forward to meet the challenges posed by the latest defensive strategies.

\section{Methodology}

\subsection{Preliminaries and Problem Formulation}

We begin by establishing the foundational concepts for adversarial attacks on standard classifiers and those employing Dummy Classes-based defenses. Consider an input \(\mathbf{x}\) with true label \(y\). A model \(f_{\boldsymbol{\theta}}\), parameterized by \(\boldsymbol{\theta}\), produces an output logit vector \(\mathbf{z} = f_{\boldsymbol{\theta}}(\mathbf{x})\). Subsequently, we sort the elements of \( \mathbf{z} \) in descending order, with \( \mathbf{z}_{\pi_1} \) representing the maximum value. 

\noindent \textbf{Standard Classifiers.} For a standard classifier tackling a \(K\)-class problem, the logit vector \(\mathbf{z}\) has \(K\) dimensions. The predicted class \(\hat{y}\) is determined by:
\begin{equation}
\hat{y} = \arg\max_{i=1,\dots,K} \mathbf{z}_i.
\end{equation}
The model correctly classifies the input if the logit of the true class \(y\) is the largest. Conversely, an adversarial attack is considered successful if it perturbs \(\mathbf{x}\) to \(\mathbf{x}'\) such that this condition is violated. A common objective for achieving this is to minimize the margin between the true class and the most competitive other class:
\begin{equation}
\mathbf{z}_y - \max_{i \neq y} \mathbf{z}_i < 0.
\label{eq:normal_target}
\end{equation}
This objective, formalized in ~\Cref{eq:normal_target}, is the cornerstone of powerful attacks like AutoAttack. The attacker's goal is to reduce this margin from a positive value (correct classification) to a negative one (misclassification).

\noindent \textbf{Dummy Classes-based Defenses.} Defenses like~\cite{wang2024new} introduce a fundamental shift by incorporating \emph{virtual} or \emph{dummy} classes. For a \(K\)-class problem, the model's output logit vector \(\mathbf{z}\) is expanded to \(2K\) dimensions. The first \(K\) logits (\(\mathbf{z}_1, ..., \mathbf{z}_K\)) correspond to the authentic classes, while the subsequent \(K\) logits (\(\mathbf{z}_{K+1}, ..., \mathbf{z}_{2K}\)) correspond to the dummy classes. Critically, each authentic class \(y\) is paired with a dedicated dummy class at index \(y+K\).

The prediction rule is modified accordingly. Let \(g(\mathbf{z}) = \arg\max_{i=1,\dots,2K} \mathbf{z}_i\). The final predicted class is then given by:
\begin{equation}
\hat{y} = \begin{cases}
g(\mathbf{z}), & \text{if } 1 \leq g(\mathbf{z}) \leq K \\
g(\mathbf{z}) - K, & \text{if } K+1 \leq g(\mathbf{z}) \leq 2K
\end{cases}.
\end{equation}
This rule means that an input can be assigned to either an authentic class or its paired dummy class. A successful adversarial attack must now cause the model to output an \emph{authentic class that is different from the true label} \(y\). This occurs if the maximum logit among all classes \emph{except} the true authentic class \(y\) and its paired dummy class \(y+K\) exceeds the maximum of these two:
\begin{equation}
\max (\mathbf{z}_y, \mathbf{z}_{y+K}) - \max_{i \notin \{y, y+K\}} \mathbf{z}_i < 0.
\label{eq:dummy_target}
\end{equation}

The discrepancy between the success criteria for standard classifiers (~\Cref{eq:normal_target}) and Dummy Classes-based defenses (~\Cref{eq:dummy_target}) reveals why traditional attacks like AutoAttack fail. An attack can successfully minimize \(\mathbf{z}_y - \max_{i \neq y} \mathbf{z}_i\) by simply increasing \(\mathbf{z}_{y+K}\), causing the input to be classified into the dummy class \(y+K\). While this satisfies ~\Cref{eq:normal_target}, it does not satisfy ~\Cref{eq:dummy_target}, as the prediction falls into the "safe sink" of the dummy class rather than a wrong authentic class. This leads to a significant overestimation of the defense's robustness. Our proposed \textbf{DAWA} is designed to address this precise limitation by formulating an optimization objective that directly targets the core mechanism of the dummy class defense.

\subsection{The \attack{} Attack Formulation}

To effectively break the Dummy Classes-based defense, an attack must simultaneously accomplish two objectives: 1) reduce the model's confidence in the true authentic class \(y\), and 2) reduce its confidence in the paired dummy class \(y+K\), thereby pushing the prediction towards a different, incorrect authentic class. We first reformulate the condition for a successful attack from ~\Cref{eq:dummy_target}:
\begin{equation}
\begin{split}
& \max (\mathbf{z}_y, \mathbf{z}_{y+K}) - \max_{i \notin \{y, y+K\}} \mathbf{z}_i = \\
&\alpha \cdot (\mathbf{z}_y - \max_{i \notin \{y, y+K\}} \mathbf{z}_i) + (1-\alpha) \cdot (\mathbf{z}_{y+K} - \max_{i \notin \{y, y+K\}} \mathbf{z}_i),
\label{eq:loss_decomposed}
\end{split}
\end{equation}
where \(\alpha\) is an indicator that selects between the two terms based on which is larger:
\begin{equation}
\alpha = \begin{cases}
1, & \text{if } \mathbf{z}_y - \mathbf{z}_{y+K} > 0 \\
0, & \text{if } \mathbf{z}_y - \mathbf{z}_{y+K} \leq 0
\end{cases}.
\label{eq:alpha_original}
\end{equation}

It can be observed that \Cref{eq:loss_decomposed} need simultaneously reduces the values of both $\mathbf{z}_y  - \max\limits_{i \neq \{y, y+K\}} \mathbf{z}_i$ and $\mathbf{z}_{y+K} - \max\limits_{i \neq \{y, y+K\}} \mathbf{z}_i$.
Inspired by the effective surrogate loss in~\cite{lafeat}, which enables efficient adversarial example generation by mitigating the negative impact of relative errors in gradient computations caused by floating-point operations, we implement the following constraints in our strategy:

To achieve the objective of reducing the value of $\mathbf{z}_y  - \max\limits_{i \neq \{y, y+K\}} \mathbf{z}_i$, we impose the following constraint:
\begin{equation}
 \mathcal{L}^{\text{ce}}\left( \frac{t^* \mathbf{z}}{(\mathbf{z}_{\pi_1}- \mathbf{z}_{\pi_2})_{\text{detach}}}, y \right)
\end{equation}

Similarly, to achieve the objective of reducing the value of $\mathbf{z}_{y+K}  - \max\limits_{i \neq \{y, y+K\}} \mathbf{z}_i$, we employ the constraint:
\begin{equation}
\mathcal{L}^{\text{ce}}\left( \frac{t^* \mathbf{z}}{(\mathbf{z}_{\pi_1}- \mathbf{z}_{\pi_2})_{\text{detach}}}, y+K \right)
\end{equation}

In our subsequent experiments, we set the hyperparameter \( t^* = 1 \), following the implementation in MIFPE. This choice is motivated by the theoretical analysis in~\cite{yu2025theoretical}, which demonstrates that \( t^* = 1 \) is very close to the theoretical optimum, thereby providing a mathematical justification for its strong performance.

\noindent \textbf{Adaptive Weighting}
As shown in \Cref{eq:alpha_original}, the value of $\alpha$ is solely determined by the sign of $(\mathbf{z}_y - \mathbf{z}_{y+K})$, which restricts $\alpha$ to binary values of 0 or 1. This limitation allows only one objective to be constrained at a time. Moreover, when $\alpha$ is confined to 0 or 1, it fails to account for the magnitude of $(\mathbf{z}_y - \mathbf{z}_{y+K})$, thereby lacking sensitivity to the actual difference between these logits. To enable simultaneous constraint of both objectives, we apply a smoothing operation to the computation of $\alpha$, defined as follows:

\begin{equation}
\alpha = \frac{1}{1 + e^{-c \cdot (\mathbf{z}_y - \mathbf{z}_{y+K})_{\text{detach}}}},
\label{eq:alpha_smooth}
\end{equation}
where $c \geq 0$ is a hyperparameter that controls the smoothness of the transition. The detachment operation applied to $(\mathbf{z}_y - \mathbf{z}_{y+K})_{\text{detach}}$ ensures that gradients are truncated from this term, thereby treating $c$ as a gradient-free constant during backpropagation.


\noindent \textbf{The Final \attack{} Loss}
We formulate the objective functions for both untargeted and targeted variants of \attack{} as follows:
\begin{equation}
\begin{split}
& \mathcal{L}^{\text{\attack}}\left( \mathbf{z}, y, K,c \right) = \\
&\alpha*\mathcal{L}^{\text{ce}}\left( \frac{t^* \mathbf{z}}{(\mathbf{z}_{\pi_1}- \mathbf{z}_{\pi_2})_{\text{detach}}}, y \right)+\\
&(1-\alpha)*\mathcal{L}^{\text{ce}}\left( \frac{t^* \mathbf{z}}{(\mathbf{z}_{\pi_1}- \mathbf{z}_{\pi_2})_{\text{detach}}}, y+K \right),
\label{method:loss}
\end{split}
\end{equation}

\begin{equation}
\begin{split}
& \mathcal{L}^{\text{\attackmt}}\left( \mathbf{z}, y_{t}, K,c \right) = \\
-&\alpha*\mathcal{L}^{\text{ce}}\left( \frac{t^* \mathbf{z}}{(\mathbf{z}_{\pi_1}- \mathbf{z}_{\pi_2})_{\text{detach}}}, y_{t} \right)-\\
&(1-\alpha)*\mathcal{L}^{\text{ce}}\left( \frac{t^* \mathbf{z}}{(\mathbf{z}_{\pi_1}- \mathbf{z}_{\pi_2})_{\text{detach}}}, y_{t}+K \right),
\label{method:targeted_loss}
\end{split}
\end{equation}
where $y_t$ is a predefined class target, $t\in \left\{ 1,2,...,K \right\} \ $ and $y_t\ne y$.

This loss function directly incentivizes the attack to find perturbations that simultaneously undermine the model's confidence in both the true class and its paired dummy class, effectively closing the "safe sink" and forcing a misclassification into another authentic class. The adaptive weighting ensures the attack dynamically focuses its effort on the more prominent of the two "safe" options during the optimization process.

\begin{algorithm}[t]
\caption{%
    The \attack{} white-box robust evaluation for dummy classes-based defences.
}\label{alg:overview}
\algnewcommand{\IfThen}[2]{
    \State \algorithmicif\ {#1}\ \algorithmicthen\ {#2}}
\algnewcommand{\IfThenElse}[3]{
    \State \algorithmicif\ {#1}\ %
    \algorithmicthen\ {#2}\ \algorithmicelse\ {#3}}
\newcommand{\algcmt}{\algorithmiccomment}
\newcommand{\submodelset}{\f}
\begin{algorithmic}[1]
    \Function{\tt\attack\_\,Attack}{$
        \submodelset,
        \x, \y, K, c, \momentum, \epsilon, I
    $}
        \State{$
            \xadv_0 \gets \project\left(
                \x + \mathbf{u}
            \right),\,\text{where}\,
            \mathbf{u} \sim \uniform{-\epsilon, \epsilon}
        $}
        \algcmt{Random init}
        \State{$\m_0 \gets 0$}
        \For{$i \in \bracks{0:I-1}$}
            \State{$
                \mathbf{z} \gets f(\xadv_i)$}
            \State{$\mathbf{v} \gets \max (\mathbf{z}_y, \mathbf{z}_{y+K}) - \max\limits_{i \neq \{y, y+K\}} \mathbf{z}_i$}
            \IfThen{$\mathbf{v} < 0$}{\Return{$\xadv_i$}}
            \algcmt{Successful attack}
            \State{$
                \bm{g}_{i + 1} \gets
                    \sign\parens{\nabla_{\xadv_i}
                        \attackloss\parens{
                       \mathbf{z},  y,K,c
                    }
                }
            $}
            \algcmt{Sign-gradient with the \attack{} loss}
            \State{$
                \stepsize \gets \epsilon \parens*{
                    1+\cos \parens*{ \nicefrac{i\pi}{I} }
                }
            $}
            \algcmt{Cosine step-size schedule}
            \State{$
                \m_{i + 1} \gets \project\left(
                    \m_i + \stepsize \bm{g}_{i + 1}
                \right)
            $}
            \algcmt{Iterative update}
            \State $\xadv_{i + 1} \gets \project\left( \xadv_i + \Delta_i \right)$
            \State \quad where $\Delta_i = \momentum ( \m_{i+1} - \xadv_i ) + (1 - \momentum) ( \xadv_i - \xadv_{i - 1} )$
            \algcmt{With momentum}
        \EndFor{}
        \State{\Return{$\xadv_I$}}
        \algcmt{Give up after $I$ iterations}
    \EndFunction%
\end{algorithmic}
\end{algorithm}%

While the new \( \attackloss \) loss
is highly effective against
dummy class based defenses we test in this paper,
we strive for further advances
in \attack's ability
to generate faster and better adversarial examples.
Inspired by publications~\cite{ye2022aaa,dong18momentum,croce20aa,lafeat},
we borrow ideas
from related adversarial attack tactics,
which includes
adopting a cosine step-size schedule~\cite{ye2022aaa},
momentum~\cite{dong18momentum,croce20aa},
and multiple target attackss~\cite{croce20aa,tramer2020adaptive,lafeat}.
We provide the overall algorithm
in \Cref{alg:overview},
which computes
an adversarial image \( \xadv_I \)
as return,
by taking as input
natural image \( \x \),
ground truth label \( \y \),
class number \(K\),
\( c \) controls the smooth of the \(\alpha\),
momentum \( \mu = 0.75 \),
\( \epsilon \) perturbation bound,
and finally the maximum number of iterations \( I \).

\section{Experimental Results}\label{sec:results}

To ensure a fair and comprehensive evaluation of our proposed \attack{} against existing attack methods, we conducted extensive experiments across different threat models and datasets. Specifically, we evaluated the \textlnorm{\infty} bounded perturbation with $\epsilon = 8/255$ on both the \cifarx{} and \cifarc{} datasets. 

For each evaluation, we compared our proposed \attack{} against several baseline attacks: standard PGD, C\&W~\cite{carlini17}, MIFPE (all using 100 steps), and the established AutoAttack benchmark~\cite{croce20aa} (using 4.9k steps). We also evaluated an extended version of our method, \attackmt{}, with 1,000 steps. All evaluations were performed on models from \cite{wang2024new} trained with three different dummy class-based defenses: PGD-AT+DUCAT, MART+DUCAT, and Consistency-AT+DUCAT.

To maintain consistency across comparisons, we kept all experimental settings identical. Firstly, we used 100 iterations for all 100-step attacks. Secondly, we adopted the same step-size schedule with a cosine decay~\cite{lafeat}, which updates the perturbation boundary $\epsilon$ at each iteration $i$ according to the formula $2\epsilon (1 + cos(\pi i/I))$,  where $I$ denotes the total number of iterations. Thirdly, we employed momentum-based updates~\cite{croce20aa} with a momentum factor of \( \momentum = 0.75 \) for all tests. Lastly, we saved the best adversarial examples generated during each attack to ensure optimal performance.

To demonstrate the effectiveness of \attack{} in achieving state-of-the-art attack success rates with only 100 iterations, and \attackmt{} with 1,000 iterations, we compare our results against both the standard AutoAttack benchmark, providing a comprehensive assessment of the true robustness of dummy class-based defenses. 
The full comparison results can be found in \Cref{tab:compare:cifarx}. We provide our key observations below.

\begin{table*}[ht]
\centering%
\footnotesize
\caption{%
    Robust accuracy (\%) comparison on CIFAR-10 and CIFAR-100 with $\epsilon = 8/255$. We evaluate models from \cite{wang2024new} trained with PGD-AT+DUCAT, MART+DUCAT, and Consistency-AT+DUCAT defenses against various attacks: standard PGD, C\&W, and MIFPE (all 100-step), our proposed \attack{} (100-step), AutoAttack (4.9k steps), and our proposed \attackmt{} (1k steps). The ``$\bm\Delta$'' column quantifies accuracy overestimation between self-reported/reproduced AutoAttack and \attackmt{}.
}\label{tab:compare:cifarx}
\adjustbox{width=\textwidth}{%
\begin{tabular}{c|c|cccc|cc|c}
\toprule
\tbox{\textbf{Defense} \\ Complexity}
    & \tbox{\textbf{Clean} \\ 1}
    & \tbox{\textbf{PGD} \\ 100}
    & \tbox{\textbf{C\&W} \\ 100}
    & \tbox{\textbf{MIFPE} \\ 100}
    & \tbox{\textbf{\attack} \\100}
    & \tbox{\textbf{AA} \\4.9k} 
    & \tbox{\textbf{\attackmt} \\1.0k}
    & \( \bm\Delta \) \\
\midrule
\multicolumn{9}{c}{%
    \textbf{\cifarx} }\\
\midrule
\multirow{1}{*}{PGD-AT+DUCAT}
    & 88.81 
        & 62.71 &  73.03 & 65.14  &  32.01
        &  58.61  
        & \textbf{ 29.52} &  \tbgreen{29.09} \\
\midrule
\multirow{1}{*}{MART+DUCAT}
    & 80.65 
        & 57.81 & 58.23 &  56.76 &  40.65
        & 50.18 
        & \textbf{39.83} & \tbgreen{10.35} \\
\midrule
\multirow{1}{*}{Consistency-AT+DUCAT}
    & 89.51 
        & 63.80 &  74.90 &  66.59 &  34.04
        &  57.18 
        & \textbf{ 31.32} & \tbgreen{25.86} \\
\midrule
\multicolumn{9}{c}{%
    \textbf{\cifarc} }\\
\midrule
\multirow{1}{*}{PGD-AT+DUCAT}
    & 70.71 
        & 29.56 &  48.60 & 28.54  &  16.56
        &  25.20  
        & \textbf{ 10.55} &  \tbgreen{14.65} \\
\midrule
\multirow{1}{*}{MART+DUCAT}
    & 56.73 
        & 34.32 & 29.66 &  29.44 &  20.92
        & 27.44 
        & \textbf{17.33} & \tbgreen{10.11} \\
\midrule
\multirow{1}{*}{Consistency-AT+DUCAT}
    & 72.29 
        & 30.73 &   49.95 &  30.30 &  19.16
        &  25.66 
        & \textbf{ 14.21} & \tbgreen{11.45} \\
\bottomrule
\end{tabular}
}\end{table*}

\textbf{AutoAttack Severely Overestimates the Robustness of Dummy Classes-based Defenses.}
Our comprehensive evaluation reveals a systematic overestimation of robustness by traditional attack benchmarks. As shown in \Cref{tab:compare:cifarx}, Dummy Classes-based defenses exhibit what appears to be state-of-the-art robustness when evaluated by AutoAttack (4.9k steps). For instance, on PGD-AT+DUCAT with CIFAR-10 under $\epsilon = 8/255$ $L_\infty$ perturbation, the defense achieves 58.61\% robust accuracy under AutoAttack evaluation, which would conventionally be interpreted as strong robustness. However, this perception is fundamentally misleading. The same defense, when evaluated with our proposed \textbf{DAWA (100 steps)}, shows a drastic reduction to 32.01\% robust accuracy, and further drops to 29.52\% with \textbf{DAWA\textsuperscript{mt} (1k steps)}—representing an absolute reduction of nearly 30\%. This pattern is consistent across all defense variants and datasets: on MART+DUCAT (CIFAR-10), robustness drops from 50.18\% (AA) to 40.65\% (DAWA) to 39.83\% (DAWA\textsuperscript{mt}); on Consistency-AT+DUCAT (CIFAR-10), from 57.18\% to 34.04\% to 31.32\%. The overestimation is even more pronounced on CIFAR-100, where PGD-AT+DUCAT drops from 25.20\% (AA) to 16.56\% (DAWA) to 10.55\% (DAWA\textsuperscript{mt}). These results unequivocally demonstrate that AutoAttack and similar traditional evaluation strategies provide a dangerously inflated assessment of these defenses' true robustness.

\textbf{DAWA Effectively Breaks Through the "Safe Sink" Mechanism with Superior Efficiency.}
Our proposed DAWA attack successfully circumvents the core defense mechanism by simultaneously optimizing against both the true class and the dummy class during adversarial example generation. The effectiveness of this approach is evidenced by DAWA achieving superior attack performance with only 100 iterations compared to AutoAttack's 4,900 iterations across all tested configurations. Notably, on PGD-AT+DUCAT (CIFAR-10), DAWA (100 steps) reduces robust accuracy by 26.6 percentage points relative to AutoAttack, while DAWA\textsuperscript{mt} (1k steps) achieves an additional 2.49 percentage point improvement. This demonstrates that DAWA not only breaks the defense more effectively but does so with dramatically improved computational efficiency—requiring less than 2\% of AutoAttack's iteration count to achieve substantially better results. The consistent performance gap across all three defense variants (PGD-AT+DUCAT, MART+DUCAT, Consistency-AT+DUCAT) and both datasets validates the generalizability of our approach.

\textbf{Traditional Attacks Fundamentally Fail Against the Dummy Class Paradigm.}
Our experiments clearly demonstrate the inherent limitations of conventional attack methodologies. Standard PGD (100 steps), C\&W (100 steps), and MIFPE (100 steps) all fail to effectively challenge these defenses, achieving robust accuracy measurements of 62.71\%, 73.03\%, and 65.14\% respectively on PGD-AT+DUCAT (CIFAR-10)—significantly higher than DAWA's 32.01\%. This performance gap stems from their shared fundamental limitation: these attacks optimize solely for misleading the true class label, an objective that Dummy Classes-based defenses are explicitly designed to counter by providing a "safe sink" for such adversarial examples. Even AutoAttack, despite its extensive 4,900 iterations and ensemble strategy, cannot overcome this architectural limitation because its constituent attacks all share the same flawed objective function. This reveals a critical weakness in the current adversarial evaluation ecosystem—defenses can achieve apparent robustness not through genuine security improvements, but by strategically exploiting known limitations in evaluation methodologies.

\begin{figure}[ht]
    \centering%
    \includegraphics[width=\linewidth, trim=0 10pt 0 0]{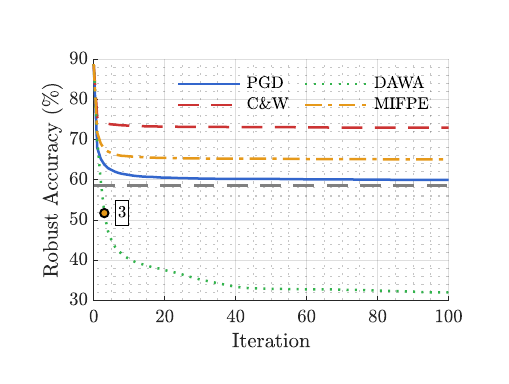}
    \caption{%
        Convergence speed comparison of four attacks (PGD, C\&W, MIFPE, and DAWA) against PGD-AT+DUCAT defense ~\cite{wang2024new} on CIFAR-10. The gray dashed line represents the AutoAttack robust accuracy (58.61\%). The number annotation on the DAWA curve indicates the iteration (3) at which DAWA first surpasses AutoAttack's performance, demonstrating its faster convergence.
        }\label{fig:results:convergence}
\end{figure}

\textbf{DAWA Achieves Rapid Convergence with State-of-the-Art Attack Performance.}
As illustrated in \Cref{fig:results:convergence}, DAWA demonstrates remarkably fast convergence compared to traditional attacks. Our method typically exceeds AutoAttack's final attack success rate within just 3 iterations, despite AutoAttack requiring 4,900 iterations to complete. This accelerated convergence stems from DAWA's direct optimization against the defense's core mechanism, avoiding the inefficient search patterns of traditional methods that inadvertently steer adversarial examples toward the dummy class. The additional improvements achieved by DAWA\textsuperscript{mt} with 1,000 iterations further demonstrate our method's ability to progressively refine attacks and uncover deeper vulnerabilities. This combination of rapid convergence and superior final performance positions DAWA as both an effective evaluation tool for researchers and a practical stress-test for defense developers seeking to develop genuinely robust models.

\textbf{The Dummy Class Mechanism Creates a False Sense of Security.}
Our results conclusively show that the high robustness reported by Dummy Classes-based defenses under traditional evaluation creates a misleading perception of security. The defense appears robust only because conventional attacks are optimized for an objective that the defense explicitly counteracts. When evaluated with an appropriate methodology like DAWA that understands and directly counteracts the dummy class mechanism, the actual vulnerability of these defenses becomes starkly apparent. This discrepancy highlights the urgent need for evolving evaluation strategies to keep pace with increasingly sophisticated defense designs, ensuring that reported robustness metrics accurately reflect real-world security guarantees rather than artifacts of evaluation methodology mismatches.

\begin{figure}[ht]
    \centering%
    \includegraphics[width=\linewidth, trim=0 10pt 0 0]{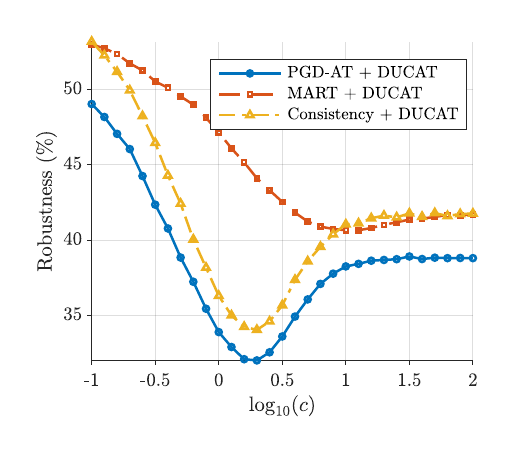}
    \caption{%
        Ablation study on the hyperparameter \( c \) (controlling \(\alpha\) smoothing in the loss function) within the DAWA evaluation strategy. The plot illustrates the untargeted attack robustness (measured after 100 iterations) as a function of \( \log_{10}(c) \) ranging from \(10^{-1.0}\) to \(10^{2.0}\), for models trained on CIFAR-10 with ResNet-18 under three strategies: PGD-AT + DUCAT, MART + DUCAT, and Consistency + DUCAT.
        }\label{fig:ablation:c}
\end{figure}

We further conduct an ablation study to analyze the impact of hyperparameter $c$ on DAWA's attack performance, where $c$ controls the smoothness of the transition function in ~\Cref{eq:alpha_smooth}. The experiment covers a range of $c$ values from $10^{-1.0}$ to $10^{2.0}$. As shown in ~\Cref{fig:ablation:c}, DAWA achieves peak attack effectiveness when $c$ falls within a moderate range. For instance, in the PGD-AT + DUCAT setting, the attack potency reaches its maximum at $c = 10^{0.3}$. However, further increasing the value of $c$ beyond this point leads to a gradual degradation in attack performance. Notably, when $c$ is set to $10^{2.0}$, the transition function reduces to a non-smooth hard threshold. Our experimental results demonstrate that introducing appropriate smooth transition through optimal $c$ values significantly enhances DAWA's attack potency compared to the non-smooth version ($c=10^{2.0}$), thereby validating the effectiveness of our proposed smoothing mechanism.


\section{Conclusion}

In this work, we have demonstrated that the current adversarial robustness evaluation paradigm faces a fundamental limitation when assessing emerging Dummy Classes-based defenses. Our analysis reveals that traditional evaluation strategies, epitomized by AutoAttack and its reliance on the $\arg\max f(\xadv) \neq y$ success criterion, systematically and severely overestimate the robustness of these defenses. This overestimation stems from a critical misalignment: while these attacks optimize for misleading the true class label, Dummy Classes-based defenses are explicitly designed to capture such attacks through a designated "safe sink", creating an illusion of robustness that does not reflect true security guarantees.

To address this critical gap, we proposed \textbf{DAWA (Dummy-Aware Weighted Attack)}, a novel evaluation methodology that fundamentally rethinks the attack objective for this class of defenses. By shifting the optimization target from the traditional $\arg\max f(\xadv) \neq y$ to our proposed $\max (\mathbf{z}_y, \mathbf{z}_{y+K}) - \max_{i \notin \{y, y+K\}} \mathbf{z}_i < 0$, DAWA simultaneously targets both the true class and the dummy class during adversarial example generation. This approach effectively neutralizes the defense's core mechanism and provides a more accurate assessment of genuine vulnerabilities.
Extensive experimental validation demonstrates DAWA's remarkable effectiveness and efficiency.

Our work underscores a broader lesson for the adversarial machine learning community: as defense strategies evolve to explicitly counter existing evaluation methodologies, our assessment tools must correspondingly advance. The significant overestimation problem we identified highlights the urgent need for continuous evolution of robustness evaluation strategies. Future defenses will likely employ increasingly sophisticated mechanisms that exploit limitations in current attack methodologies, necessitating the development of more adaptive, defense-aware evaluation frameworks. We hope that DAWA serves as both an immediate solution for accurately assessing Dummy Classes-based defenses and an inspiration for developing the next generation of robustness evaluation methodologies that can keep pace with increasingly sophisticated defense strategies.
\clearpage

{
    \small
    \bibliographystyle{ieeenat_fullname}
    \bibliography{main}
}


\end{document}